\documentclass[letterpaper]{article} % DO NOT CHANGE THIS
\usepackage{aaai24}  % DO NOT CHANGE THIS
\usepackage{times}  % DO NOT CHANGE THIS
\usepackage{helvet}  % DO NOT CHANGE THIS
\usepackage{courier}  % DO NOT CHANGE THIS
\usepackage[hyphens]{url}  % DO NOT CHANGE THIS
\usepackage{graphicx} % DO NOT CHANGE THIS
\usepackage{natbib}  % DO NOT CHANGE THIS AND DO NOT ADD ANY OPTIONS TO IT
\usepackage{caption} % DO NOT CHANGE THIS AND DO NOT ADD ANY OPTIONS TO IT
\usepackage{algorithm,algpseudocode}
\usepackage{newfloat}
\usepackage{listings}
\DeclareCaptionStyle{ruled}{labelfont=normalfont,labelsep=colon,strut=off} % DO NOT CHANGE THIS
\floatstyle{ruled}
\newfloat{listing}{tb}{lst}{}
\floatname{listing}{Listing}
\usepackage{amsmath} % used for the {aligned} equations.
\usepackage{subfig} % used for the figures of the ablation study.
\usepackage{multirow} % used for table
\usepackage[super]{nth}
\usepackage{mathtools}
\usepackage{gensymb}
\usepackage{breakcites}
\usepackage{booktabs}
\usepackage{multirow}
\usepackage{adjustbox,lipsum}
\usepackage{hyperref}

\usepackage{listings}
\usepackage{bm}
\usepackage{amssymb}
\title{Compressing Image-to-Image Translation GANs Using Local Density Structures on Their Learned Manifold}
\author{
    Alireza Ganjdanesh\equalcontrib$^{\text{1}}$, Shangqian Gao\equalcontrib$^{\text{2}}$, Hirad Alipanah$^{\text{3}}$, Heng Huang$^{\text{1}}$}
\affiliations{
    \textsuperscript{\rm 1} Department of Computer Science, University of Maryland College Park, College Park, MD 20742, USA
    \\
    \textsuperscript{\rm 2} Department of Electrical and Computer Engineering, University of Pittsburgh, Pittsburgh, PA 15261, USA\\
    \textsuperscript{\rm 3} Department of Mechanical Engineering and Materials Science, University of Pittsburgh, Pittsburgh, PA 15261, USA
    \\
    aliganj@umd.edu, shg84@pitt.edu, hia21@pitt.edu, heng@umd.edu
}

\begin{document}

\maketitle

\begin{abstract}
Generative Adversarial Networks (GANs) have shown remarkable success in modeling complex data distributions for image-to-image translation.~Still,~their high computational demands prohibit their deployment in practical scenarios like edge devices.~Existing GAN compression methods mainly rely on knowledge distillation or convolutional classifiers' pruning techniques.~Thus, they neglect the critical characteristic of GANs:~their~\textit{local density structure over their learned manifold}.~Accordingly, we approach GAN compression from a new perspective by explicitly encouraging the pruned model to preserve the density structure of the original parameter-heavy model on its learned manifold.~We facilitate this objective for the pruned model by partitioning the learned manifold of the original generator into local neighborhoods around its generated samples.~Then, we propose a novel pruning objective to regularize the pruned model to preserve the local density structure over each neighborhood, resembling the kernel density estimation method.~Also, we develop a collaborative pruning scheme in which the discriminator and generator are pruned by two pruning agents.~We design the agents to capture interactions between the generator and discriminator by exchanging their peer's feedback when determining corresponding models' architectures. Thanks to such a design, our pruning method can efficiently find performant sub-networks and can maintain the balance between the generator and discriminator more effectively compared to baselines during pruning, thereby showing more stable pruning dynamics. Our experiments on image translation GAN models, Pix2Pix and CycleGAN, with various benchmark datasets and architectures demonstrate our method's effectiveness.
\end{abstract}

%%%%%%%%%%%%%%%%%%%%%%%%%%%%%%%%%%%%%%%%%%%%%%%%%%%%%%%%%%%%%%%%%%%%%%%%%%%%%%%

\section{Introduction}

Image-to-Image translation~\cite{isola2017image,zhu2017unpaired} Generative Adversarial Networks (I2IGANs)~\cite{goodfellow2014generative} have shown excellent performance in many real-world computer vision applications:~style transfer~\cite{huang2017arbitrary}, converting a user's sketch to a real image~\cite{park2019semantic}, super resolution~\cite{ledig2017photo,wang2018esrgan}, and pose transfer~\cite{wang2018video,chan2019everybody}.~Yet, I2IGANs require excessive compute and memory resources.~Moreover, the mentioned tasks require real-time user interaction, making it infeasible to deploy I2IGANs on mobile and edge devices in Artificial Intelligence of Things (AIoT) with limited resources.~Thus, developing compression schemes for GANs to preserve their performance and reduce their computational burden is highly desirable.~As the training dynamics of GAN models are notoriously unstable, GAN compression is much more challenging than pruning other deep models like Convolutional Neural Network (CNN) classifiers.

%%%%%%%%%%%%%%%%%%%%%%%%%%%%%%%%%%%%%%%%%%%%%%%%%%%%%%%%%%%%%%%%%%%%%%%%%%%%%%%
Despite that notable efforts have been made to compress CNNs~\cite{han2015learning,sandler2018mobilenetv2,rastegari2016xnor,li2016pruning,ye2020good,wan2020fbnetv2}, GAN compression has only been explored in recent years.~Early works have proposed a combination of prominent CNN pruning techniques like Neural Architecture Search (NAS)~\cite{gong2019autogan,li2020gan,jin2021teachers,gao2020adversarialnas,li2022learning}, Knowledge Distillation~\cite{aguinaldo2019compressing,wang2020gan,chang2020tinygan,chen2020distilling,DBLP:conf/icml/FuCWLLW20,DBLP:conf/aaai/HouY0SCW21,zhang2022wavelet,zhang2022region}, and channel pruning~\cite{li2016pruning,li2022learning} to prune GANs.~However, GAN Slimming~\cite{wang2020gan} demonstrated that heuristically stacking several CNN pruning methods for GAN compression can degrade the final performance mainly due to the instabilities of GAN training.~GCC~\cite{li2021revisiting} empirically showed the importance of restricting the discriminator's capacity during compression.~It demonstrated that the previous methods' unsatisfactory performance might be because they only pruned the generator's architecture while using the full-capacity discriminator. By doing so, the adversarial game cannot maintain the Nash Equilibrium state, and the pruning process fails to converge appropriately.~Although some of these methods have shown competitive results~\cite{li2021revisiting,jin2021teachers}, they do not explicitly consider an essential characteristic of GANs as generative models during pruning,~which is their~\textit{density structure over their learned manifold.}

%%%%%%%%%%%%%%%%%%%%%%%%%%%%%%%%%%%%%%%%%%%%%%%%%%%%%%%%%%%%%%%%%%%%%%%%%%%%%%%
In this paper, we propose a novel GAN Compression method by enforcing the similarity of the density structure of the original parameter-heavy model and the pruned model over the learned manifold of the original model.~Our intuition is that the difference in density structures can serve as the supervision signal for pruning.~Specifically, at first, we partition the learned manifold of the original model into local neighborhoods.~We approximate each neighborhood with a generated sample and its nearest neighbors on the original model's manifold.~We leverage a pretrained self-supervised model fine-tuned on the training dataset to find the neighborhoods.~Then, we introduce an adversarial pruning objective to encourage the pruned model to have a similar local density structure to the original model on each neighborhood.~By doing so, we break down the task of preserving the whole density structure of the original model on its learned manifold into maintaining local density structures on neighborhoods of its manifold, which resembles kernel density estimation \cite{parzen1962estimation}. In addition, we design a new adversarial GAN compression scheme in which two pruning agents (we call them $gen_G$ and $gen_D$, which determine the structure of the generator $G$ and discriminator $D$) collaboratively play our proposed adversarial game. Specifically, each agent takes the architecture embedding of its colleague as its input to determine the structure of its corresponding model in each iteration. By doing so, $gen_G$ and $gen_D$ will be able to effectively preserve the balance between the capacities of $G$ and $D$ and keep the adversarial game close to the Nash Equilibrium state during the pruning process.~We summarize our contributions as follows:

%%%%%%%%%%%%%%%%%%%%%%%%%%%%%%%%%%%%%%%%%%%%%%%%%%%%%%%%%%%%%%%%%%%%%%%%%%%%%%%

\begin{itemize}
    \item We propose a novel GAN compression method that encourages the pruned model to have a similar local density structure as the original model on neighborhoods of the original model's learned manifold.
    
    \item We design two pruning agents that collaboratively play our adversarial pruning game to compress both the generator and discriminator together.~By doing so, our method can effectively preserve the balance between the capacities of the generator and discriminator and show more stable pruning dynamics while outperforming baselines.
    \item Our extensive experiments on Pix2Pix~\cite{isola2017image} and CycleGAN~\cite{zhu2017unpaired} on various datasets demonstrate our method's effectiveness.
\end{itemize}

\begin{figure*}[t]
\centering
\includegraphics[scale=0.153]{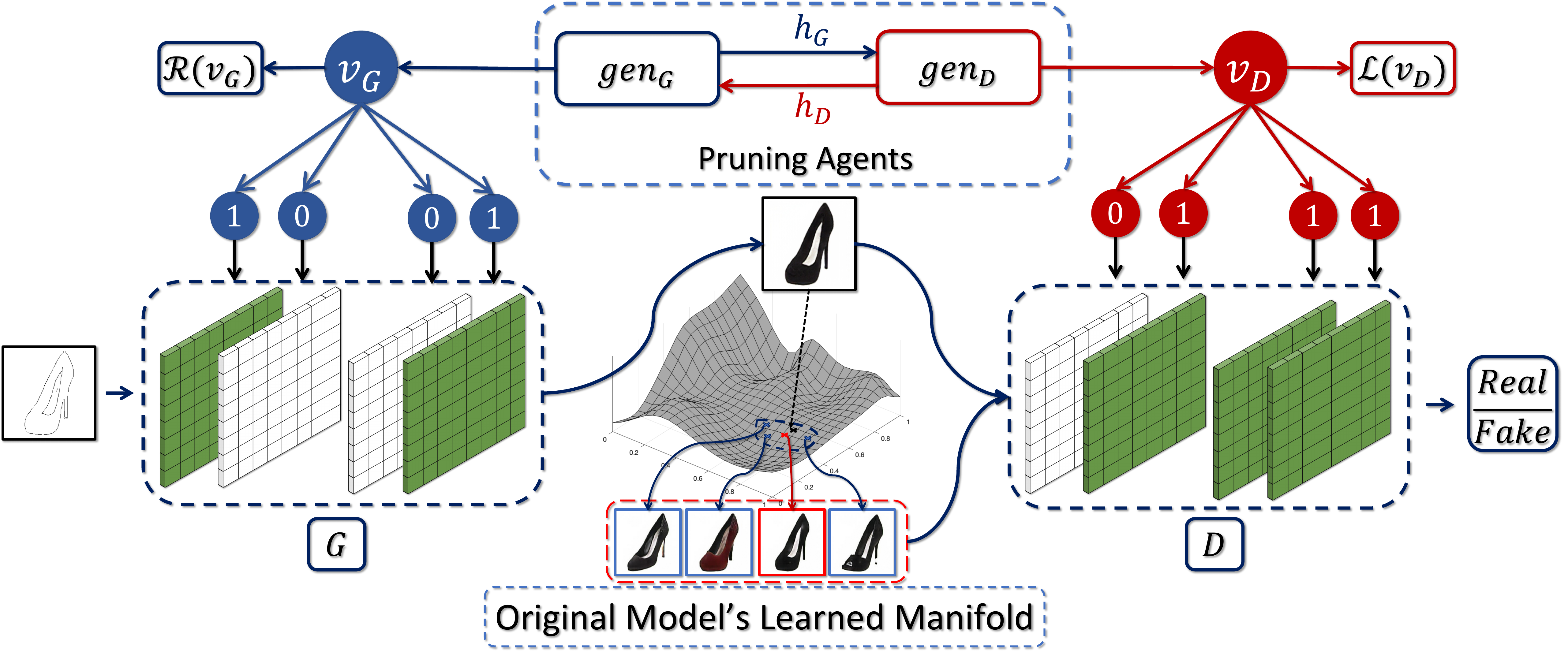}
\caption{\small Our GAN pruning method.~We encourage the pruned generator to preserve the density structure of the original model over its learned manifold during pruning.~To do so, we partition the manifold into local neighborhoods around the samples generated by the original generator (Fig.~\ref{neighborhoods}) and represent each local neighborhood with a `Center' sample (shown with a red frame) and its neighbors (blue frames).~We use these samples as `real' samples and the one generated by the pruned generator as a `fake' one in our adversarial pruning objective. We implement our adversarial game with two pruning agents, $gen_G$ and $gen_D$, that collaboratively learn to prune the original pretrained $G$ and $D$. $gen_G$ ($gen_D$) takes the architecture embedding of their colleague $gen_D$ ($gen_G$) when determining the architecture of $G$ ($D$).~By doing so, $gen_G$ and $gen_D$ can maintain the balance between the capacity of $G$ and $D$ during pruning and make the process stable. (Fig.~\ref{fig:loss_stat})}
\label{overall-scheme}
\end{figure*}

%%%%%%%%%%%%%%%%%%%%%%%%%%%%%%%%%%%%%%%%%%%%%%%%%%%%%%%%%%%%%%%%%%%%%%%%%%%%%%%

\section{Related Work}

%%%%%%%%%%%%%%%%%%%%%%%%%%%%%%%%%%%%%%%%%%%%%%%%%%%%%%%%%%%%%%%%%%%%%%%%%%%%%%%
\noindent\textbf{GAN Compression:} GANs require two orders of magnitude more computation than CNNs~\cite{li2020gan}. Hence, GAN compression is crucial prior to deploying them on edge devices.~Search-based methods~\cite{shu2019co,li2020gan,lin2021anycost,wang2021coarse} search for a lightweight architecture for the generator but are extremely costly due to their vast search space.~Pruning methods~\cite{li2020gan,jin2021teachers,wang2020gan,yu2020self} prune the redundant weights of the generator's architecture but neglect the discriminator.~They result in an unbalanced generator and discriminator capacities, leading to mode collapse~\cite{li2021revisiting}.~To address this problem, discriminator-free methods~\cite{ren2021online,DBLP:conf/icml/FuCWLLW20} distill the generator into a compressed model without using the discriminator.~In another direction, GCC~\cite{li2021revisiting} and Slimmable GAN~\cite{DBLP:conf/aaai/HouY0SCW21} prune both the generator and discriminator together.~Slimmable GAN sets the discriminator's layers' width proportional to the ones for the generator during pruning.~Yet, GCC empirically showed there is no linear relation between the number of channels of the generator and optimal discriminator, and Slimmable GAN's approach is sub-optimal. Inspired by GCC, we use two pruning agents that learn to determine the architectures of the generator and discriminator in our proposed adversarial game.~Each agent gets feedback from its peer when determining its corresponding model's architecture.~Thus, they can effectively preserve the balance between the generator and discriminator and stabilize the pruning process.

%%%%%%%%%%%%%%%%%%%%%%%%%%%%%%%%%%%%%%%%%%%%%%%%%%%%%%%%%%%%%%%%%%%%%%%%%%%%%%%

\noindent\textbf{Manifold Learning for GANs:}~The manifold hypothesis indicates that high-dimensional data like natural images lie on a nonlinear manifold with much smaller intrinsic dimensionality~\cite{tenenbaum2000global}.~Accordingly, a group of methods alter the training~\cite{casanova2021instanceconditioned} and/or architecture of GANs by using several generators~\cite{khayatkhoei2018disconnected}, including a manifold learning step in the discriminator~\cite{ni2022manifold}, and local coordinate coding based sampling~\cite{cao2018adversarial}.~Yet, one cannot use these methods for pruning GANs that are pretrained with other methods.~Another group of ideas observed that semantically meaningful paths and neighborhoods exist in the latent space of trained GANs.~\cite{oldfield2021tca,harkonen2020ganspace}.~Inspired by these methods, we propose to partition the learned manifold of a pretrained GAN into overlapping neighborhoods.~Then, we develop an adversarial pruning scheme that encourages the pruned model to have a similar density structure to the original one over each neighborhood.

%%%%%%%%%%%%%%%%%%%%%%%%%%%%%%%%%%%%%%%%%%%%%%%%%%%%%%%%%%%%%%%%%%%%%%%%%%%%%%%

\noindent\textbf{Network Pruning:}~Model compression~\cite{ghimire2022survey} is a well-studied topic, and proposed methods have primarily focused on compressing CNN classifiers.~They use various techniques like weight pruning~\cite{han2015learning}, light architecture design~\cite{tan2019efficientnet,howard2017mobilenets}, weight quantization~\cite{rastegari2016xnor}, structural pruning~\cite{li2016pruning,ye2020good,ganjdanesh2022interpretations,he2018amc}, knowledge distillation~\cite{DBLP:conf/nips/BaC14}, and NAS~\cite{wu2019fbnet,ganjdanesh2023effconv}. We focus on developing a compression method for GANs, which is a more challenging task due to the instability and complexity of their training~\cite{wang2020gan}.

%%%%%%%%%%%%%%%%%%%%%%%%%%%%%%%%%%%%%%%%%%%%%%%%%%%%%%%%%%%%%%%%%%%%%%%%%%%%%%%
\section{Proposed Method}
% \subsection{Overview}

We develop a new GAN compression method that explicitly regularizes the pruned model not to forget the density structure of the original one over its learned manifold. However, directly applying such regularization along with compression objectives can make the pruning process unstable because of the complex nature of training GANs.~Thus, we simplify this objective for the pruned generator in two steps.~At first, we propose to partition the learned manifold of the original model into local neighborhoods, each consisting of a sample and its nearest neighbors on the manifold.~We employ a self-supervised model fine-tuned on the training dataset to approximately find such neighborhoods.~Then, we propose an adversarial pruning objective to enforce the pruned model to have a density structure similar to the original model over each local neighborhood.~Finally, we introduce a novel GAN compression scheme in which we use two pruning agents - called $gen_G$ and $gen_D$ - that determine the architectures of the generator $G$ and discriminator $D$, respectively.~$gen_G$ and $gen_D$ play the adversarial game introduced in the previous step in a collaborative manner to find the optimal structure of $G$ and $D$. In each step,~$gen_G~(gen_D)$ gets feedback from its peer~$gen_D~(gen_G)$ about the architecture of~$D~(G)$ when determining the architecture of $G~(D)$. By doing so, they can maintain the balance between the generator and discriminator during pruning and stabilize the pruning process.~We show our pruning scheme in Fig.~\ref{overall-scheme}.

%%%%%%%%%%%%%%%%%%%%%%%%%%%%%%%%%%%%%%%%%%%%%%%%%%%%%%%%%%%%%%%%%%%%%%%%%%%%%%%
\begin{figure*}[!t]
\centering
\includegraphics[scale=0.154]{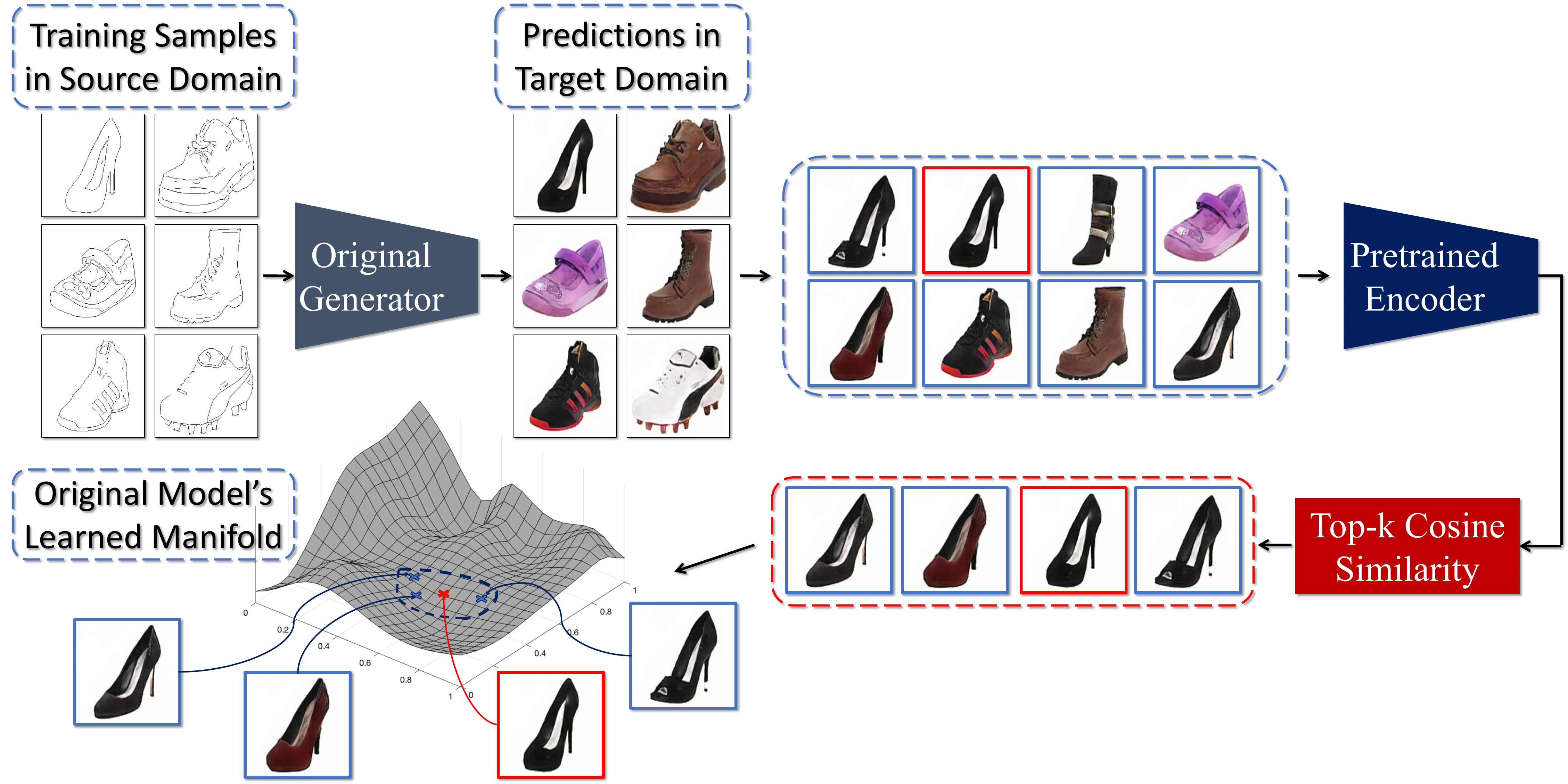}
\caption{\small Our method to find local neighborhoods on the learned manifold of the original generator.~\textbf{(Top Left):}~First, we obtain the original model's predictions in the target domain for training samples in the source domain.~\textbf{(Right and Down):}~We call the sample that we want to find its local neighborhood on the manifold `Center' sample (shown with Red solid frame).~We pass the predicted samples in the previous step to a pretrained self-supervised encoder~\cite{caron2020unsupervised} that is fine-tuned on the target images in the training dataset.~Then, we take the samples whose representations have the highest cosine similarity with the representation of the `Center' sample as its approximate neighbors on the manifold.~Neighbor samples and the approximate neighborhood on the manifold are shown with blue crosses and a dashed line.}

\label{neighborhoods}
\end{figure*}

%%%%%%%%%%%%%%%%%%%%%%%%%%%%%%%%%%%%%%%%%%%%%%%%%%%%%%%%%%%%%%%%%%%%%%%%%%%%%%%
\subsection{Notations}

We denote an I2IGAN model's source and target domains with $\mathcal{X}$ and $\mathcal{Y}$, respectively.~We show the training dataset as $\mathcal{D}=\{\{(x_{i})\}_{i=1}^{N}$, $\{(y_{j})\}_{j=1}^{M}\}$ such that $(x,y) \sim \mathcal{P}(\textbf{x}, \textbf{y})$ where $\mathcal{P}$ is the underlying joint distribution over the source and target domains.~$N$ and $M$ can be equal for the paired datasets~\cite{isola2017image} or be unequal for unpaired models~\cite{zhu2017unpaired}.~We represent the generator and discriminator with $G$ and $D$.~Also, we denote pruning agents determining the architectures of $G$ and $D$ during pruning with $gen_G(\cdot)$ and $gen_D(\cdot)$. The goal of the generator is to learn the distribution of corresponding $y\in\mathcal{Y}$ in the target domain conditioned on a sample $x\in\mathcal{X}$ from the source domain. We denote the learned manifold of the original generator in the domain $\mathcal{Y}$ with $\mathcal{M}_y$.

%%%%%%%%%%%%%%%%%%%%%%%%%%%%%%%%%%%%%%%%%%%%%%%%%%%%%%%%%%%%%%%%%%%%%%%%%%%%%%%
\subsection{Finding Local Neighborhoods on the Learned Data Manifold of the Original Generator}\label{finding-local-neighbor}
As mentioned above, our idea is to guide the pruning process of a GAN model by regularizing the pruned generator to have a similar density structure to the original model over its learned manifold, $\mathcal{M}_y$.~To simplify this objective for the pruned model, we partition $\mathcal{M}_y$ into local neighborhoods $\mathcal{N}_{y_i}$ containing a sample ${y_i}$ and its nearest neighbors on the manifold $\mathcal{M}_y$.~The intuition is that separately modeling the density structure over each neighborhood is easier than modeling all of them simultaneously, which resembles the kernel density estimation method~\cite{parzen1962estimation}.

To find the local neighborhoods, on the one hand, we get inspirations from recent works that observed that latent spaces of GAN models have semantically meaningful paths and neighborhoods~\cite{tzelepis2021warpedganspace,oldfield2021tca,shen2021closed,choi2022do,voynov2020unsupervised}. On the other hand, self-supervised pretrained encoders~\cite{caron2020unsupervised,chen2020simple,chen2020improved,grill2020bootstrap} have shown significant results in unsupervised clustering and finding semantically similar samples without using labels on the manifold of their input data.~Accordingly, at first, we fine-tune an encoder pretrained by SwAV~\cite{caron2020unsupervised} on the samples $\{(y_{j})\}_{j=1}^{M}$ in the training dataset.~Then, we use the training dataset and pass the training samples $x_i$ into the original generator to obtain its predicted samples $\mathcal{D}'_y=\{y'_i\}_{i=1}^{N}$ on $\mathcal{M}_y$. Finally, we approximately model the neighborhood of $y'_i$ over $\mathcal{M}_y$ by finding its $k$ nearest neighbor samples in $\mathcal{D}'_y$ using our fine-tuned encoder.~Formally, given a fine-tuned self-supervised encoder $E$, we find $k$ nearest neighbors of $y'_i$ in $\mathcal{D}'_y$, denoted by $\mathcal{N}_{y_i, k}$, as follows: (Fig.~\ref{neighborhoods})

\begin{equation}~\label{eq-neighbor}
\begin{small}
    \begin{aligned}
    \mathcal{E'} &= \{e'_j\}_{j=1}^{N},~e'_j = E(y'_j)\\
    Sim_i &= \{Cosine(e'_i, e'_j)\}_{\substack{j=1,\ j\neq i}}^{N}\\
    Cosine(e'_i, e'_j) &= ||e_i^{'T} e'_j||/(||e'_i||||e'_j||)\\
    \mathcal{N}_{y'_i, k} &= \{y'_j | e'_j \in \text{Top-}k(Sim_i)\}
    \end{aligned}
\end{small}
\end{equation}

\noindent\textit{i.e.,} we take samples in $\mathcal{D}'_y$ that their encoded representations have the highest cosine similarity with the one for $y'_i$ as its neighbors on $\mathcal{M}_y$.~We use the cosine similarity metric as it has been widely used in the nearest neighbor classifiers~\cite{vinyals2016matching,touvron2021grafit} and self-supervised learning~\cite{caron2020unsupervised}.

%%%%%%%%%%%%%%%%%%%%%%%%%%%%%%%%%%%%%%%%%%%%%%%%%%%%%%%%%%%%%%%%%%%%%%%%%%%%%%%
\subsection{Pruning}
In this section, first, we elaborate on our pruning objective. Then, we introduce our pruning agents.

% \subsubsection{Pruning Objective:}

\noindent\textbf{Pruning Objective:}
We regularize the pruned generator to have a similar density structure to the original one over each local neighborhood of the learned manifold of the original model ($\mathcal{M}_y$). Formally, we approximately represent the neighborhood around each sample $y'_i$ with samples in $\mathcal{N}_{y'_i, k}$ in Eq.~\ref{eq-neighbor}.~Then, we define our adversarial training objective to regularize the pruned model to preserve the local density structure of the original model in each neighborhood:

\begin{equation}\label{obj-conditional}
% \begin{small}
\small
\begin{aligned}
    &\min_{\theta_G}\max_{\theta_D} \mathbb{E}_{(x, y') \sim \mathcal{P'}(\textbf{x, y})}[\mathbb{E}_{y''\sim \mathcal{N}_{y'}}[f_D(D(x, y'';v_D))]]+ \\ 
    &\mathbb{E}_{x\sim \mathcal{P}(\textbf{x})}\mathbb{E}_{\xi}[f_G(D(x, G(x, \xi;v_G);v_D))] + \lambda_1\mathcal{R}(v_G) - \lambda_2\mathcal{L}(v_D)
\end{aligned}
% \end{small}
\end{equation}

\noindent$\mathcal{P'}(y)$ is the marginal distribution of the original generator model on $\mathcal{Y}$.~$G$ and $D$ are the pretrained generator and discriminator.~$\theta_G$ and $\theta_D$ are parameters of $gen_G$ and $gen_D$.~$v_G$ and $v_D$ are architecture vectors that determine the structures of $G$ and $D$. They are functions of $\theta_G$ and $\theta_D$.~$f_D$ and $f_G$ are GAN objectives that are least squares for CycleGAN~\cite{zhu2017unpaired} and hinge loss for Pix2Pix~\cite{isola2017image}. $\xi$ represents the randomness in the generator implemented with Dropout~\cite{srivastava2014dropout} for Pix2Pix and Cycle GAN.~$\mathcal{R}$ is the regularization term to enforce the desired compression ratio on the generator, and $\mathcal{L}$ imposes sparsity on the architecture of the discriminator.~Penalizing unimportant components for discriminator is crucial to maintain the capacity balance between $G$ and $D$ during pruning and keeping them close to the Nash Equilibrium state, as pointed out by GCC~\cite{li2021revisiting}.~Moreover, our objective is similar to the Kernel Density Estimation (KDE)~\cite{parzen1962estimation} method, which aims to model the local density around each data point.~Yet, in contrast with KDE, we use an implicit objective to encourage the pruned model to have a similar density structure to the original one in our method. In addition, in Obj.~\ref{obj-conditional}, the parameters of $G$ and $D$ are inherited from the original pretrained models, and we do not train parameters of $G/D$ along with $\theta_{G}/\theta_{D}$ to prevent instability.~We note that our pruning objective does not require paired images during pruning, even for GAN models such that use a paired dataset for training. The reason is that our objective employs the samples generated by the original generator to approximate its density structure over each local neighborhood. Thus, it can readily prune both paired~\cite{isola2017image} and unpaired conditional GANs~\cite{zhu2017unpaired}.

%%%%%%%%%%%%%%%%%%%%%%%%%%%%%%%%%%%%%%%%%%%%%%%%%%%%%%%%%%%%%%%%%%%%%%%%%%%%%%%
% \subsubsection{Pruning Agents} 
\noindent\textbf{Pruning Agents:} Inspired by GCC~\cite{li2021revisiting} that demonstrated that preserving the balance between the capacity of $G$ and $D$ is crucial for pruning stability, we prune both $G$ and $D$ during the pruning process.~To do so, we introduce a novel GAN compression scheme in which we use two pruning agents, $gen_G$ and $gen_D$, to predict $1)$ architecture vectors and $2)$ architecture embeddings for $G$ and $D$. The former is a binary vector determining the architecture of the corresponding model ($G/D$), and the latter is a compact representation describing its state (architecture). To preserve the balance between the capacity of $G$ and $D$, we design our pruning scheme such that each pruning agent $gen_G/gen_D$ considers the architecture embedding of $D/G$ when determining the architecture of $G/D$.~We implement pruning agents with a GRU~\cite{DBLP:conf/emnlp/ChoMGBBSB14} model and dense layers (more details in supplementary materials) and take the outputs of the GRU units of the pruning agents $gen_G/gen_D$ as their corresponding model's ($G/D$) architecture embedding, summarizing the information about its architecture.

In each step of the adversarial game, $gen_G$ and $gen_D$ exchange their architecture embeddings.~Then, each of them determines its corresponding model's architecture while knowing the other model's structure (architecture embedding), making the pruning process stable and efficient.~(Fig.~\ref{overall-scheme}) We denote the architecture vector determining the architecture of $G/D$ with $v_G/v_D \in \{0, 1\}$.~As these vectors have discrete values, their gradients \emph{w.r.t} agents' parameters cannot be calculated directly.~Thus, we use Straight-through Estimator (STE)~\cite{bengio2013estimating} and Gumbel-Sigmoid reparametrization trick~\cite{DBLP:conf/iclr/JangGP17} to calculate the gradients. The sub-network vector $v_G$ is calculated as:
% Formally, the generation of sub-network vector $v_G$ can be represented by:
\begin{equation}~\label{eq-vG}
    % \begin{small}
    \begin{aligned}
    v_G &= \text{round}(\text{sigmoid}(-(o_G+g)/\tau)), \\ o_G, h_G &= gen_G(h_D; \theta_G)
    \end{aligned}
    % \end{small}
\end{equation}
where round$(\cdot)$ rounds input to its nearest integer, sigmoid$(\cdot)$ is the sigmoid function, $\tau$ is the temperature parameter to control the smoothness, and $g\in \text{Gumbel}(0,1)$ is a random vector sampled from the Gumbel distribution~\cite{gumbel1954statistical}. $h_D$ is the architecture embedding for $D$, which is the input for $gen_G$.~$o_G$ and $h_G$ are the output of $gen_G$ and its architecture embedding for $G$. Similarly, $v_D$ is calculated as:

\begin{equation}~\label{eq-vD}
    \begin{aligned}
    v_D &= \text{round}(\text{sigmoid}(-(o_D+g)/\tau)), \\ o_D, h_D &= gen_D(h_G; \theta_D)
    \end{aligned}
\end{equation}
\noindent The calculation from $o_*$ to $v_*$ ($*\in\{G, D\}$) can be seen as using straight-through Gumbel-Sigmoid~\cite{DBLP:conf/iclr/JangGP17} to approximate sampling from the Bernoulli distribution.~We provide our pruning algorithm and details of the calculation of $o_*$ and $h_*$ in the supplementary.

Predicted architecture vectors $v_G$ and $v_D$ determine the architectures of $G$ and $D$.~For the generator $G$, we aim to reduce MACs to reach a given budget. To do so, we use the following regularization objective:
\begin{equation}~\label{eq-MACs}
    \mathcal{R}(v_G) =  \log(\max(T(v_G) ,pT_{\text{total}}) / pT_{\text{total}}),
\end{equation}
where $p$ is the predefined threshold for pruning, $T(v_G)$ is the MACs of the current sub-network chosen by $v_G$, and $T_{\text{total}}$ is the total prunable MACs. 

Different from $G$, it is not obvious how to set a specific computation budget for $D$, as shown in GCC~\cite{li2021revisiting}. Instead of using a predefined threshold, we encourage $gen_D$ to automatically identify the sub-network that can keep the Nash Equilibrium given the $v_G$. Inspired by the functional modularity~\cite{csordas2021are}, we add a penalty to $v_D$ to discover the suitable sub-module (sub-network) to keep the Nash Equilibrium:
\begin{equation}~\label{eq-FM}
    \mathcal{L}(v_D) = \sum v_D/ |v_D|,
\end{equation}
where $|v_D|$ is the number of elements in $v_D$.~The goal of Eq.~\ref{eq-FM} is to penalize unimportant elements in $v_D$, so the remaining elements can maintain Nash Equilibrium given $v_G$.

%%%%%%%%%%%%%%%%%%%%%%%%%%%%%%%%%%%%%%%%%%%%%%%%%%%%%%%%%%%%%%%%%%%%%%%%%%%%%%%
% \subsubsection{Finetuning} 
\noindent\textbf{Finetuning:} After the pruning stage, we employ the trained $gen_G/gen_D$ from the pruning process to prune $G$/$D$ according to their predictions, $v_G$/$v_D$. Then, we finetune $G$ and $D$ together with the original objectives of their GAN methods~\cite{isola2017image,zhu2017unpaired}. Similar to GCC~\cite{li2021revisiting}, we also apply knowledge distillation (KD) for finetuning, but we show in our ablation experiments that our model can achieve high performance even without KD.

%%%%%%%%%%%%%%%%%%%%%%%%%%%%%%%%%%%%%%%%%%%%%%%%%%%%%%%%%%%%%%%%%%%%%%%%%%%%%%%

\section{Experiments}

% \subsection{Setup}
We perform our experiments with Pix2Pix and Cycle-GAN models. For all experiments, we set $\lambda_1=3.0$, $\lambda_2=0.1$, and the number of neighbor samples $k=5$ during pruning.~We also find that our model is not very sensitive to the choice of $\lambda_1$ and $\lambda_2$. (more details in ablation experiments) We set the number of pruning epochs to $10\%$ of the original model's training epochs.~We refer to supplementary materials for more details of our experimental setup.

\begin{table*}[t!]
\centering
\caption{ Quantitative comparison of our proposed method with state-of-the-art GAN compression methods.}
\resizebox{0.81\linewidth}{!}{
\begin{tabular}{ccc|c|c|c|c}
    \hline
    \multirow{2}{*}{Model} & \multirow{2}{*}{Dataset} & \multirow{2}{*}{Method} & \multirow{2}{*}{MACs} & \multirow{2}{*}{Compression Ratio} & \multicolumn{2}{c}{Metric} \\ \cline{6-7} & & & & & \multicolumn{1}{c|}{FID ($\downarrow)$} & mIoU ($\uparrow$) \\ \hline
    \multirow{12}{*}{Pix2Pix} & \multirow{8}{*}{Cityscapes} & Original~\cite{isola2017image} & 18.60 G & - & - & 42.71  \\
                        & & GAN Compression~\cite{li2020gan}& $5.66$~G & $69.57\%$ & - & $40.77$ \\
                        & & CF-GAN~\cite{wang2021coarse}& $5.62 $~G & $69.78\%$ & - & $42.24$  \\
                        & & CAT~\cite{jin2021teachers}& $5.57 $~G & $70.05\%$ & - & $42.53$  \\
                        & & DMAD~\cite{li2022learning}& $3.96 $~G & $78.71\%$ & - & $40.53$  \\
                        & & WKD~\cite{zhang2022wavelet}& $3.88$~G & $79.13\%$ & - & $42.93$ \\
                        & & RAKD~\cite{zhang2022region}& $3.88$~G & $79.13\%$ & - & $42.81$ \\
                        & & Norm Pruning~\cite{li2016pruning,liu2017learning}& $3.09$~G & $83.39\%$ & - & $38.12$  \\
                        & & GCC~\cite{li2021revisiting}& $3.09$~G & $83.39\%$ & - & $42.88$  \\
                        & & \textbf{MGGC (Ours)} & $3.05$~G & $83.60\%$ & - & $\textbf{44.63}$ \\ \cline{2-7} & 
    \multirow{4}{*}{Edges2Shoes} & Original~\cite{isola2017image} & $18.60$ G & - & 34.31 & - \\
                        & & Pix2Pix 0.5$\times$~\cite{isola2017image} & $4.65$~G & $75.00\%$ & 52.02 & - \\
                        & & CIL~\cite{kim2022cut} & $4.57$~G & $75.43\%$ & 44.40 & - \\
                        & & DMAD~\cite{li2022learning} & $2.99$~G & $83.92\%$ & 46.95 & - \\
                        & & WKD~\cite{zhang2022wavelet} & $1.56$~G & $91.61\%$ & 80.13 & - \\
                        & & RAKD~\cite{zhang2022region} & $1.56$~G & $91.61\%$ & 77.69 & - \\
                        & & \textbf{MGGC (Ours)} & $2.94$~G & $84.19\%$ & \textbf{42.02} & - \\ \hline
    \multirow{16}{*}{CycleGAN} & \multirow{12}{*}{Horse2Zebra}  & Original~\cite{zhu2017unpaired} & $56.80$~G & - & 61.53 & - \\
                        & & Co-Evolution~\cite{shu2019co} & $13.40$~G & $76.41\%$ & 96.15 & - \\
                        & & GAN Slimming~\cite{wang2020gan} & $11.25$~G & $80.19\%$ & 86.09 & - \\
                        & & AutoGAN-Distiller~\cite{DBLP:conf/icml/FuCWLLW20} & $6.39$~G & $88.75\%$ & 83.60 & - \\
                        & & WKD~\cite{zhang2022wavelet} & $3.35$~G & $94.10\%$ & 77.04 & - \\
                        & & RAKD~\cite{zhang2022region} & $3.35$~G & $94.10\%$ & 71.21 & - \\
                        & & GAN Compression~\cite{li2020gan} & $2.67$~G & $95.30\%$ & 64.95 & - \\
                        & & CF-GAN~\cite{wang2021coarse} & $2.65$~G & $95.33\%$ & 62.31 & - \\
                        & & CAT~\cite{jin2021teachers} & $2.55$~G & $95.51\%$ & 60.18 & -      \\
                        & & DMAD~\cite{li2022learning} & $2.41$~G & $95.76\%$ & 62.96 & -      \\
                        & & Norm Prune~\cite{li2016pruning,liu2017learning} & $2.40$~G & $95.77\%$ & 145.1 & -\\
                        & & GCC~\cite{li2021revisiting} & $2.40$~G & $95.77\%$ & 59.31 & - \\
                        & & \textbf{MGGC (Ours)} & $2.50$~G & $95.60\%$ & \textbf{55.06} & - \\ \cline{2-7} & 
    \multirow{6}{*}{Summer2Winter} & Original~\cite{zhu2017unpaired} & 56.80 G & - & 79.12 & - \\
                        & & Co-Evolution~\cite{shu2019co} & $11.06$~G & $80.53\%$ & 78.58 & - \\
                        & & AutoGAN-Distiller~\cite{DBLP:conf/icml/FuCWLLW20} & $4.34$~G & $92.36\%$ & 78.33 & - \\
                        & & DMAD~\cite{li2022learning} & $3.18$~G & $94.40\%$ & 78.24 & - \\
                        & & \textbf{MGGC (Ours)} & $1.69$~G & $97.02\%$ & \textbf{77.33} & - \\
                        & & \textbf{MGGC (Ours)} & $2.97$~G & $94.77\%$ & \textbf{75.85} & -     \\
    \hline
\end{tabular}
}
\label{table:comparison}
\end{table*}

%%%%%%%%%%%%%%%%%%%%%%%%%%%%%%%%%%%%%%%%%%%%%%%%%%%%%%%%%%%%%%%%%%%%%%%%%%%%%%%

\subsection{Results}
% \subsubsection{Comparison with State of the Art Methods}
\noindent\textbf{Comparison with State of the Art Methods:} We summarize the quantitative results of our method and baselines in Tab.~\ref{table:comparison}.~As can be seen, our method, \textbf{MGGC} (\textbf{M}anifold \textbf{G}uided \textbf{G}AN \textbf{C}ompression), can achieve the best performance~\emph{vs.}~computation rate trade-off compared to baselines in all experiments.~For Pix2Pix on Cityscapes, MGGC reduces MACs by $83.60\%$, achieving the highest MACs compression rate, and improves mIoU by $1.92$ compared to the original model, significantly outperforming baselines. On Edges$2$Shoes, MGGC prunes $0.27\%$ more MACs than DMAD~\cite{li2022learning} while outperforming it with a large margin of $4.93$ FID.~Although WKD~\cite{zhang2022wavelet} and RAKD~\cite{zhang2022region} achieve higher compression ratio, their final model has significantly worse FID. For CycleGAN on Horse$2$Zebra, MGGC shows a pruning ratio very close ($95.60\%$~\emph{vs.}~$95.77\%$) to GCC~\cite{li2021revisiting} and accomplishes $55.06$ FID, significantly outperforming GCC by $4.25$ FID. On Summer$2$Winter, MGGC attains $94.77\%$ MACs compression rate, slightly more than DMAD with $94.40\%$, and shows $2.39$ less FID. Remarkably, it can outperform other baselines even with an extreme MACs compression rate of $97.02\%$, and yet, reaching $77.33$ FID. In summary, our results demonstrate the effectiveness of our method that explicitly focuses on the differences between the density structure of the pruned model and the original one over its learned manifold during pruning.

%%%%%%%%%%%%%%%%%%%%%%%%%%%%%%%%%%%%%%%%%%%%%%%%%%%%%%%%%%%%%%%%%%%%%%%%%%%%%%%

\begin{figure}[hbt!]
\centering
\includegraphics[scale=0.09]{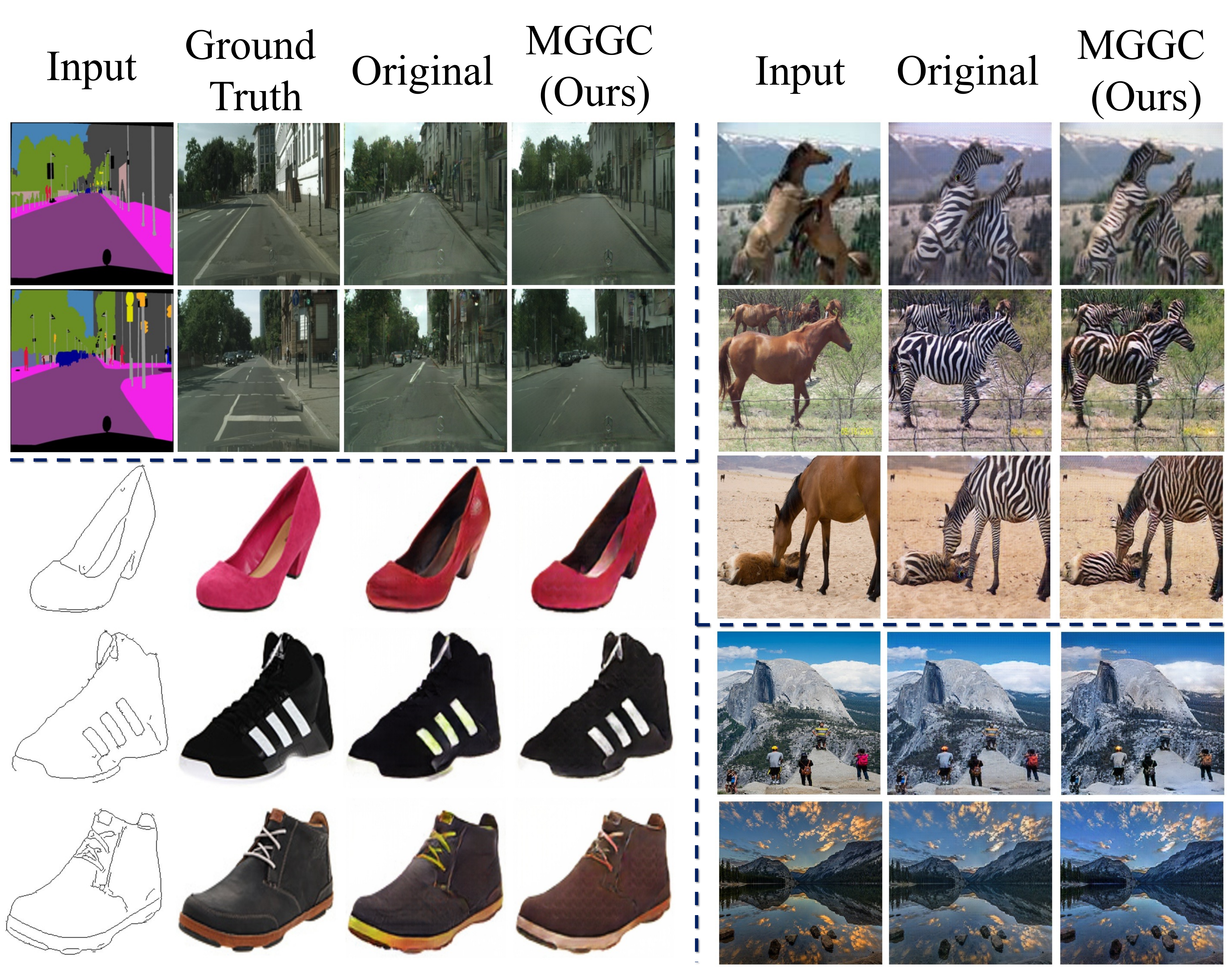}
\caption{\small Qualitative results for \textbf{1) Pix2Pix:} Cityscapes (top left), Edges$2$Shoes (bottom left), and \textbf{2)~CycleGAN:} Horse$2$Zebra (top right) and Summer$2$Winter (bottom right).}
\label{visualization}
\end{figure}

% \subsubsection{Qualitative Results}
\noindent\textbf{Qualitative Results:} We visualize predictions of our pruned models and the original ones in Fig.~\ref{visualization}. As can be seen, our pruned model can preserve the fidelity of images with a much lower computational burden.~Also,~its superior performance compared to baselines is observable in the samples for Cityscapes~(better preserving street structure) and Horse$2$Zebra~(more faithful background color).

\begin{figure*}[tbh!]
    \centering
    \subfloat[$\lambda_1=4.0$]{
	\begin{minipage}[b]{.33\linewidth}
			\centering
			\includegraphics[width=.87\textwidth]{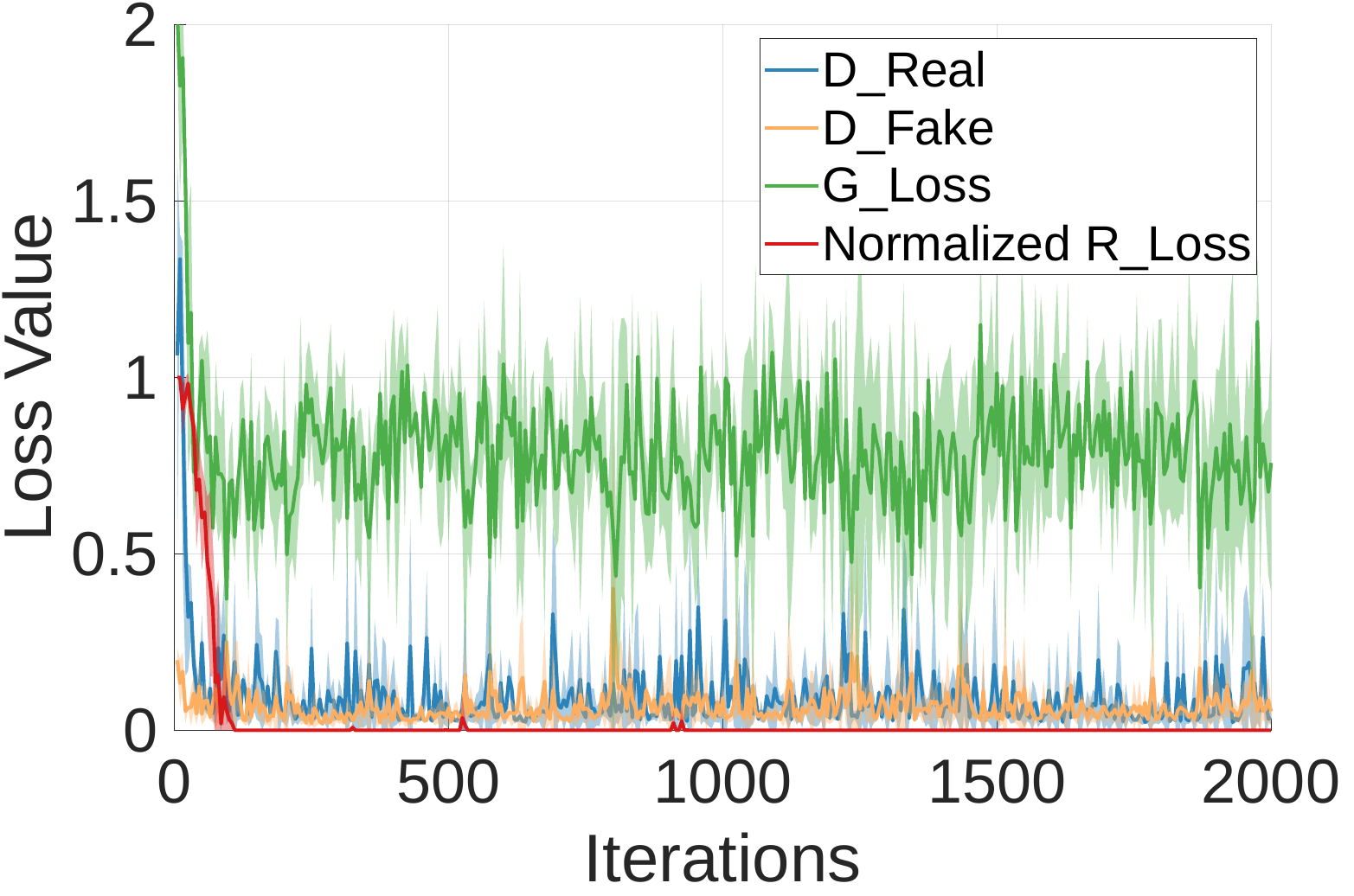}
	\end{minipage}}
	\subfloat[$\lambda_1=3.0$]{
		\begin{minipage}[b]{.33\linewidth}
			\centering
			\includegraphics[width=.87\textwidth]{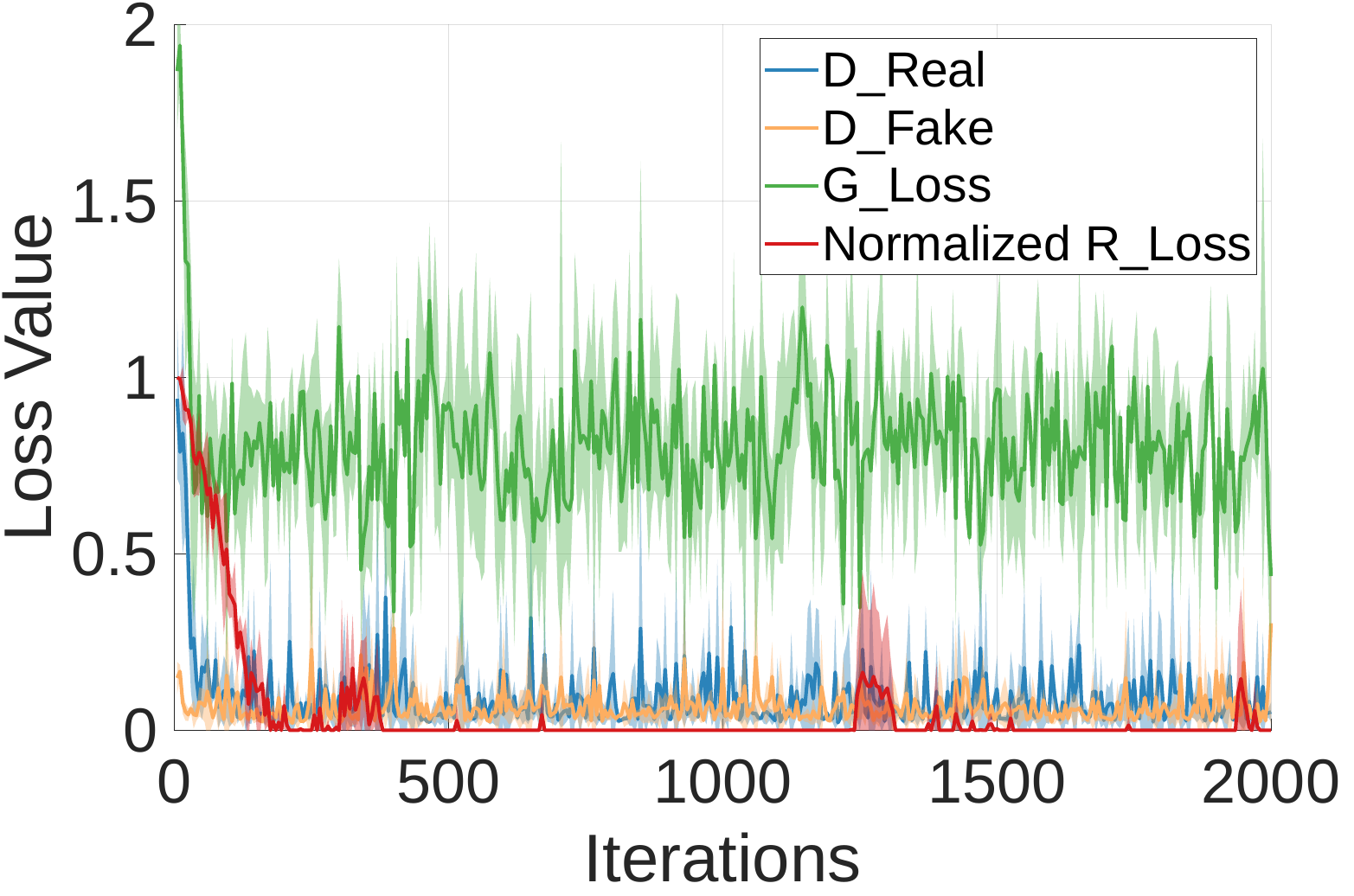}
	\end{minipage}}
	\subfloat[$\lambda_1=2.0$]{
		\begin{minipage}[b]{.33\linewidth}
			\centering
			\includegraphics[width=.87\textwidth]{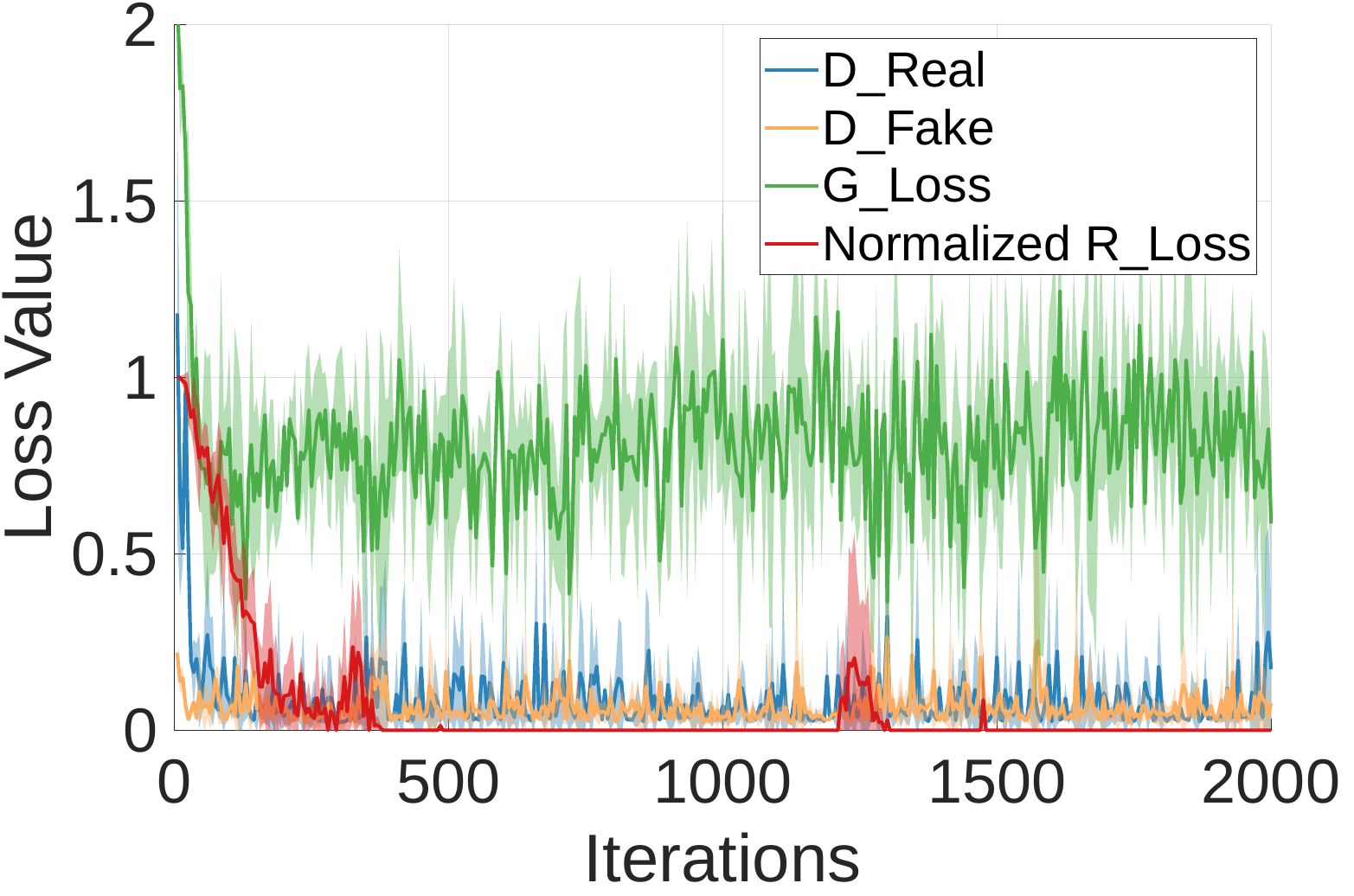}
	\end{minipage}}
	\\
	    \subfloat[$\lambda_1=4.0$]{
	\begin{minipage}[b]{.33\linewidth}
			\centering
			\includegraphics[width=.87\textwidth]{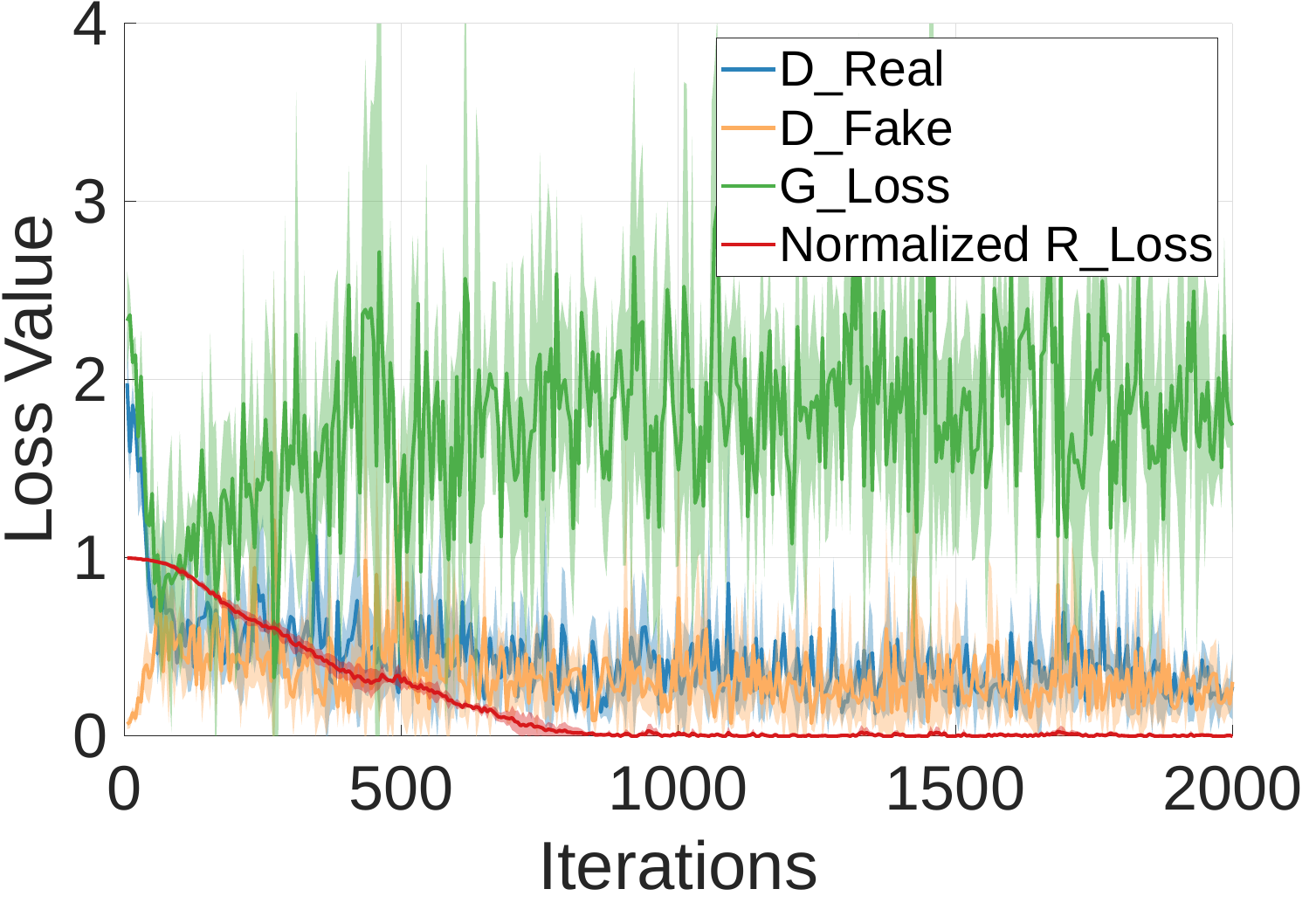}
	\end{minipage}}
	\subfloat[$\lambda_1=3.0$]{
		\begin{minipage}[b]{.33\linewidth}
			\centering
			\includegraphics[width=.87\textwidth]{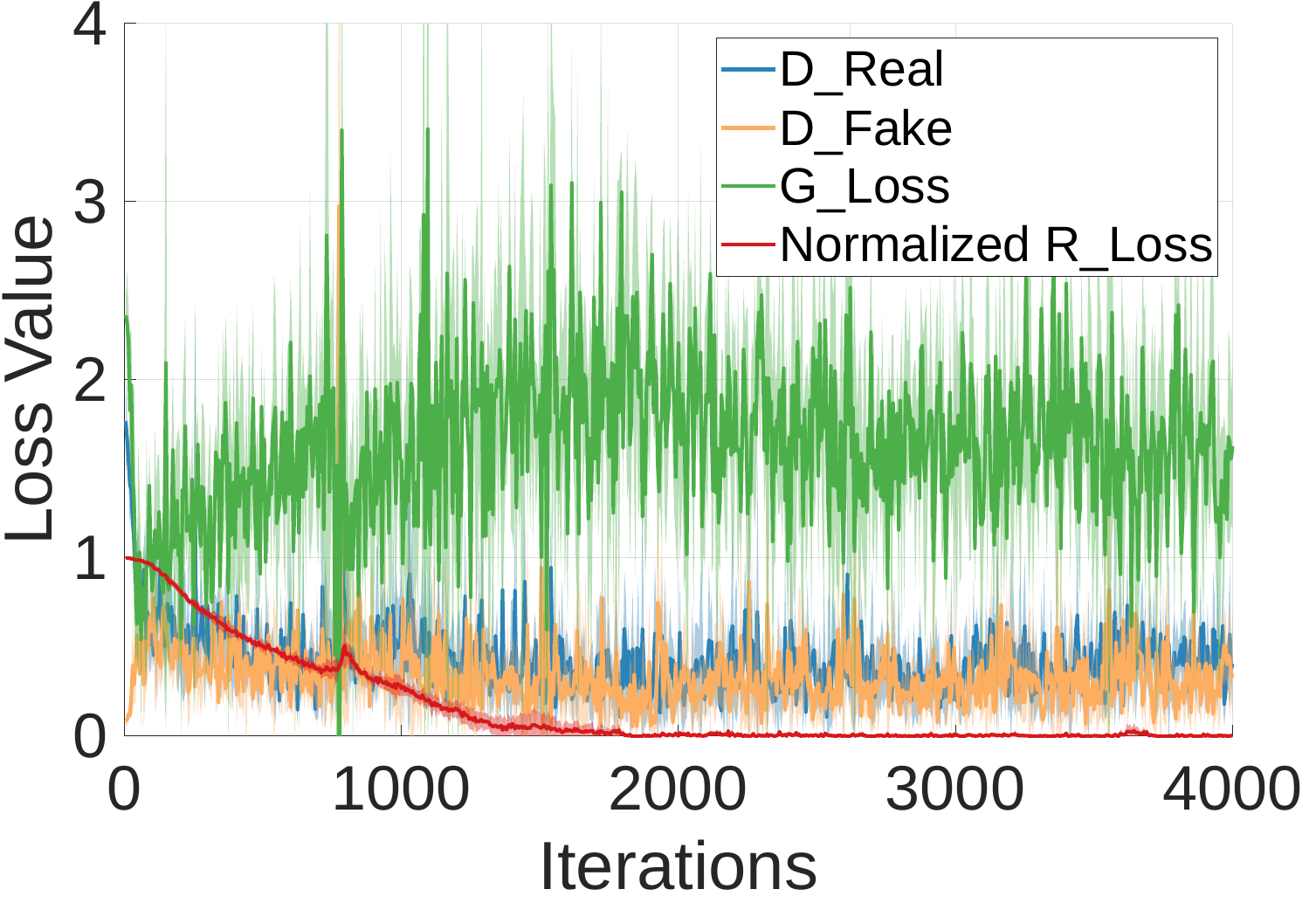}
	\end{minipage}}
	\subfloat[$\lambda_1=2.0$]{
		\begin{minipage}[b]{.33\linewidth}
			\centering
			\includegraphics[width=.87\textwidth]{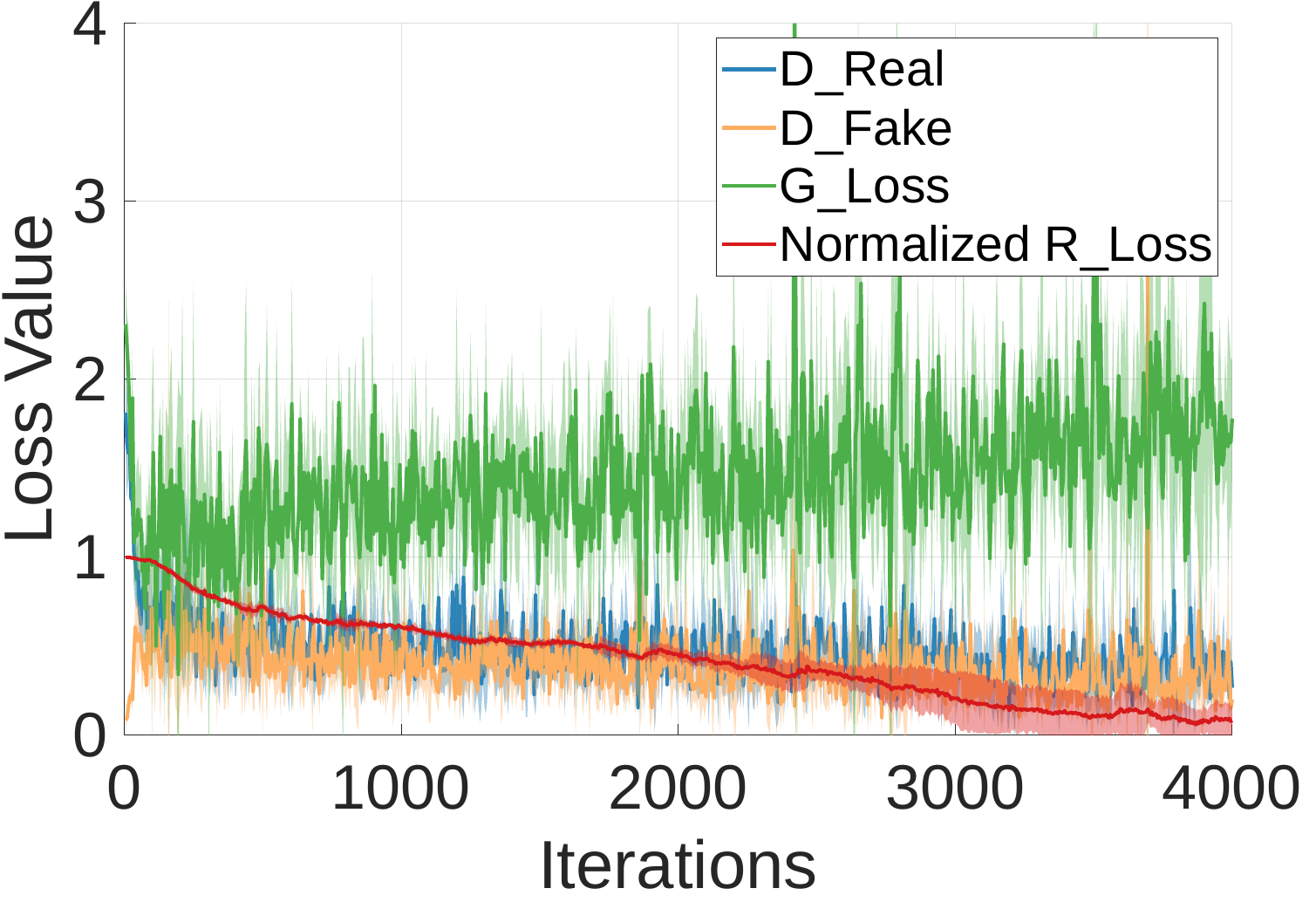}
	\end{minipage}}
    \caption{\small Different losses given different $\lambda_1$ during the pruning process. (a)-(c) Loss values for CycleGAN on Horse$2$Zebra dataset. (d)-(f) Loss values for Pix$2$Pix on Cityscapes dataset. We normalize $\mathcal{R}$ to the range $[0,1]$ for better visualization.}
    \label{fig:loss_stat}
\end{figure*}

%%%%%%%%%%%%%%%%%%%%%%%%%%%%%%%%%%%%%%%%%%%%%%%%%%%%%%%%%%%%%%%%%%%%%%%%%%%%%%%

\subsection{Stability of Our Pruning Method} \label{stability-analysis}
We explore our method's stability by visualizing different loss values and the resource loss $\mathcal{R}$ during pruning.~Loss values for CycleGAN on Horse2Zebra are shown in Fig.~\ref{fig:loss_stat} (a)-(c),~and the ones for Pix2Pix on Cityscapes are presented in Fig.~\ref{fig:loss_stat} (d)-(f).~We found in our experiments that $\lambda_2$ has less impact on pruning dynamics than $\lambda_1$,~which is expected as the generator's capacity is more crucial to GANs' performance than the discriminator's capacity.~Thus, we alter $\lambda_1$ to explore how it impacts the pruning process.~We can observe that if $\lambda_1$ be large enough~(Fig.~\ref{fig:loss_stat} (a)-(e)),~the loss values for $G$ and $D$ will get stable and remain close to each other when $\mathcal{R}$ converges to zero.~Further, the difference between loss values for G and D in Fig.~\ref{fig:loss_stat} (d)-(f) for our model is much smaller than the same value for GCC shown in GCC~\cite{li2021revisiting} Fig. 5(a), which demonstrates that our method can preserve the balance between $G$ and $D$ more effectively than GCC during pruning.~In summary, visualizations in Fig.~\ref{fig:loss_stat} show that our method can successfully preserve the balance between $G$, $D$ after achieving the desired computation budget, leading to attain competitive final performance metrics.

%%%%%%%%%%%%%%%%%%%%%%%%%%%%%%%%%%%%%%%%%%%%%%%%%%%%%%%%%%%%%%%%%%%%%%%%%%%%%%%
\subsection{Ablation Study} \label{ablation-experiments}
We present ablation results of our method with different settings in Tab.~\ref{table:ablation}.~We construct a Baseline by only pruning the generator $G$ with a naive parameterization $v_G = \text{round}(\text{sigmoid}(-(\theta +g)/\tau))$~(Eq.~\ref{eq-vG}) when using the full capacity discriminator.~The results demonstrate that using pruning agents, pruning the discriminator, and establishing a feedback connection between $gen_G$ and $gen_D$ provide substantial performance gain compared to the baseline on Pix2Pix and CycleGAN models.~This observation suggests that sophisticated designs of pruning agents are beneficial for pruning conditional GAN models.~Further, considering local density structures over neighborhoods of the learned manifold $(k>0)$ boosts the performance of our method significantly than not leveraging them $(k=0).$~Remarkably, our method can almost recover the original model's performance for Pix2Pix~($42.53$~\emph{vs.}~$42.71$) and even outperform it for CycleGAN ($58.64$~\emph{vs.}~$61.53$)~without using Knowledge Distillation (KD) in the finetuning process.~These results show that the density structure over the learned manifold contains valuable information for pruning GAN models. Utilizing KD can further improve our model.~Compared to GCC~\cite{li2021revisiting}, we do not use online distillation for KD. Also, we do not learn the discriminator's architecture during finetuning, which saves computational costs.

\begin{table}[hbt!]
\caption{Ablation results of our proposed method.}
\centering
\scalebox{0.7}{
\begin{tabular}{c|cc|cc}
\hline
\multirow{2}{*}{Settings} & \multicolumn{2}{c|}{Pix$2$Pix - Cityscapes}              & \multicolumn{2}{c}{Cycle GAN - Horse$2$Zebra}           \\ \cline{2-5} 
                          & \multicolumn{1}{c|}{mIoU ($\uparrow$)} & MACs                    & \multicolumn{1}{c|}{FID ($\downarrow$)} & MACs                    \\ \hline
Baseline                 & \multicolumn{1}{c|}{39.37}   & \multirow{6}{*}{3.05 G} & \multicolumn{1}{c|}{73.28}  & \multirow{6}{*}{2.50 G} \\
+ D pruning               & \multicolumn{1}{c|}{39.84}   &                         & \multicolumn{1}{c|}{70.41}  &                         \\
+ Pruning Agents          & \multicolumn{1}{c|}{40.68}   &                         & \multicolumn{1}{c|}{67.60}  &                         \\
+ G$\leftrightarrow$D Feedback          & \multicolumn{1}{c|}{40.99}   &                         & \multicolumn{1}{c|}{66.72}  &                         \\
+ Manifold Pruning        & \multicolumn{1}{c|}{42.53}   &                         & \multicolumn{1}{c|}{58.64}  &                         \\
+ Knowledge Distillation  & \multicolumn{1}{c|}{44.63}   &                         & \multicolumn{1}{c|}{55.06}  &                         \\ \hline
Original                  & \multicolumn{1}{c|}{42.71}   & 18.60 G                  & \multicolumn{1}{c|}{61.53}  & 56.80 G                  \\ \hline
\end{tabular}
}
\label{table:ablation}
\end{table}

%%%%%%%%%%%%%%%%%%%%%%%%%%%%%%%%%%%%%%%%%%%%%%%%%%%%%%%%%%%%%%%%%%%%%%%%%%%%%%%
\subsection{Visualization of Local Neighborhoods}
\begin{figure}[hbt!]
\centering
\includegraphics[scale=0.09]{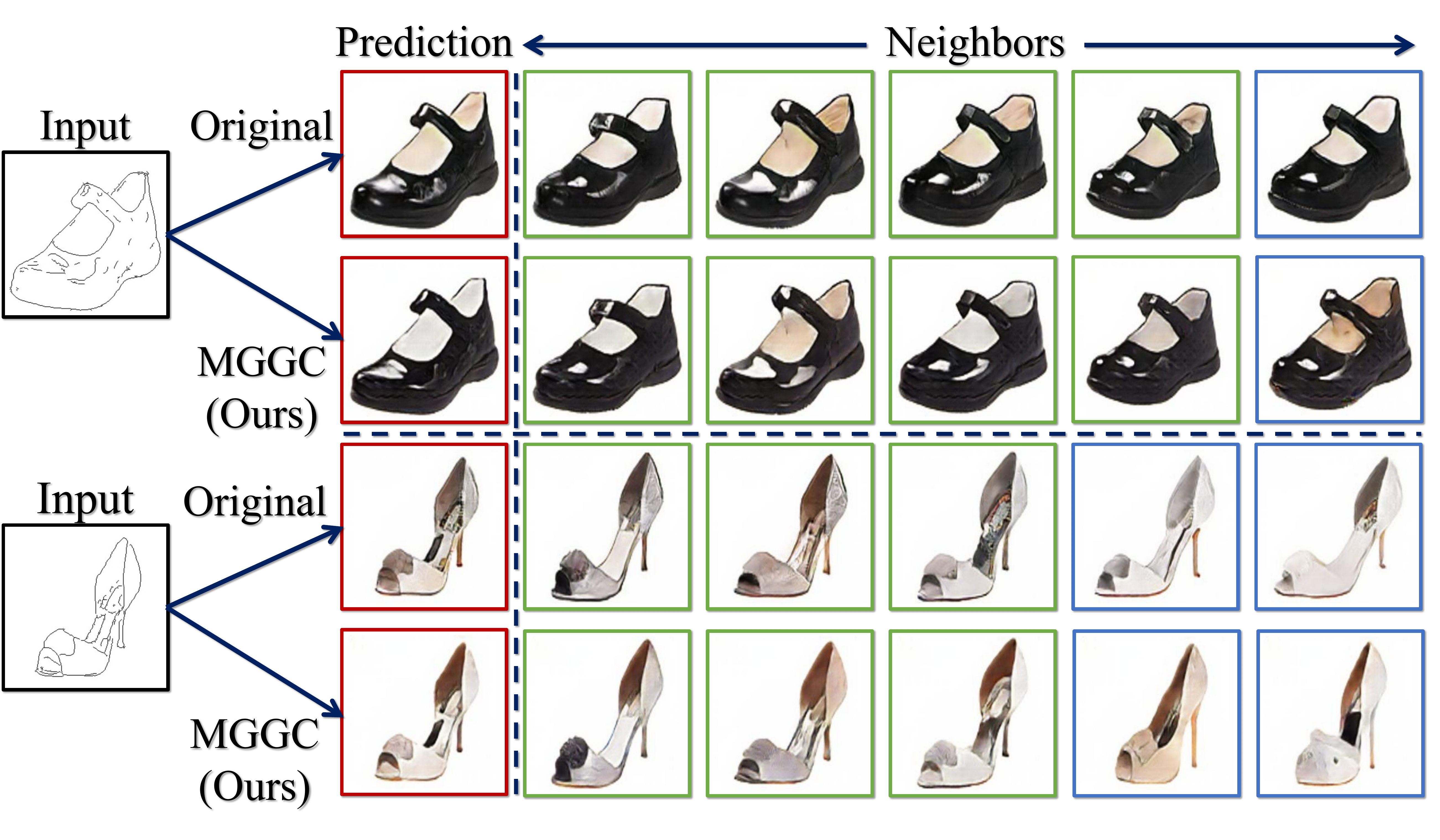}
\caption{\small Visualization of approximate neighborhoods on the learned manifold of our pruned model \emph{vs.} the original model.}
\label{local-neighbor-visualization}
\end{figure}

We explore the difference between the learned neighborhoods of our model \emph{vs.}~the original one in Fig.~\ref{local-neighbor-visualization}.~Samples with red frames are the predictions, and their neighbors on the right are obtained with the method in Eq.~\ref{eq-neighbor}. In each column, green frames show the samples having the same source domain (`Edge') image, and the blue ones mean different source domain inputs. As can be seen, most of the neighbor samples of our pruned model are identical to those for the pruned model. In addition, the samples with different source images still have a semantically meaningful connection to the predicted image. These results demonstrate that our pruning objective, Eq.~\ref{obj-conditional}, can effectively regularize the pruned model to have a similar density structure over the neighborhoods of the original model's manifold.

%%%%%%%%%%%%%%%%%%%%%%%%%%%%%%%%%%%%%%%%%%%%%%%%%%%%%%%%%%%%%%%%%%%%%%%%%%%%%%%

\section{Conclusions}

We introduced a new compression method for image translation GANs that explicitly regularizes the pruned model to have a similar density structure to the original one on the original model's learned data manifold.~We simplify this objective for the pruned model by leveraging a pretrained self-supervised encoder fine-tuned on the target dataset to find approximate local neighborhoods on the manifold.~Then, we proposed our adversarial pruning objective that motivates the pruned generator to have a similar density structure to the original model over each local neighborhood.~In addition, we proposed a novel pruning scheme that uses two pruning agents to determine the architectures of the generator and discriminator. They collaboratively play our adversarial pruning game such that in each step, each pruning agent takes the architecture embedding of its colleague as its input and determines its corresponding model's architecture.~By doing so, our agents can effectively maintain the balance between the generator and discriminator, thereby stabilizing the pruning process.~Our experimental results clearly illustrate the added value of using the learned density structure of a GAN model for pruning it compared to the baselines that directly combine CNN classifiers' compression techniques. 

\section*{Acknowledgements}
This work was partially supported by NSF IIS 2347592, 2348169, 2348159, 2347604, CNS 2347617, CCF 2348306, DBI 2405416.

\bibliography{main}

\clearpage

\section{Details of Our Experiments}
We mainly follow the experimental settings of the previous works in the literature~\cite{li2021revisiting,li2020gan,li2022learning}.~We perform our experiments on the prominent image-to-image translation methods, Pix2Pix~\cite{isola2017image} and CycleGAN~\cite{zhu2017unpaired}.~The generator model has a U-Net~\cite{ronneberger2015u} style architecture for the Pix2Pix experiments~\cite{isola2017image,li2021revisiting} and a ResNet~\cite{he2016deep} style for CycleGAN experiments~\cite{li2020gan,li2021revisiting,zhu2017unpaired}. For Pix2Pix experiments, we evaluate our model on the Edges$2$Shoes~\cite{yu2014fine} and Cityscapes~\cite{cordts2016cityscapes} datasets. For CycleGAN experiments, we use Horse2Zebra~\cite{zhu2017unpaired} and Summer2Winter~\cite{zhu2017unpaired} datasets.~We follow the evaluation metrics in the literature~\cite{li2021revisiting,li2020gan,li2022learning} to evaluate our model,~\textit{i.e}., we use mean Intersection over Union (mIoU) for the experiment for Pix2Pix on Cityscapes. We use Fr\'echet Inception Distance (FID)~\cite{heusel2017gans} as our evaluation metric for other experiments. Higher mIoU and Lower FID values mean superior generative capability. For all datasets, we set $\lambda_1=3.0$, $\lambda_2=0.1$, and the number of neighbor samples $k=5$ during pruning.~We also found that the setting $k\in\{3, 4, 5\}$ results in close final performance. Thus, we set $k=5$ in all experiments. We implemented our experiments with PyTorch and ran them on a server with 2 NVIDIA P40 GPUs.

\section{Our Pruning Agents} \label{pruning-agents}
We provide implementation details of our proposed pruning agents in this section. 
\subsection{Architectural Design}
We use a Gated Recurrent Unit (GRU)~\cite{DBLP:conf/emnlp/ChoMGBBSB14} and dense layers to implement our pruning agents. The detailed architecture is shown in Tab.~\ref{tab-arch}.~The inputs $x_*^l$ are randomly generated, and they are fixed after initialization during training and inference.

\begin{table}[hbt!]
\centering
\caption{The architecture of $gen_*$ ($*\in\{G, D\}$) used in our method.}
    {
        \begin{tabular}{c}
            \toprule
                  Inputs $x_*^l$, $l=1,\cdots, L_*$\\
            \midrule
                  GRU(128, 256), WeightNorm, ReLU\\
             %\midrule
                  $\textrm{Dense}_l$(256, $C_*^l$), WeightNorm, $l=1,\cdots, L_*$\\
             \midrule
                  Outputs ${o}_*^l$, $l=1,\cdots, L_*$; $h_*^{L_*}$\\
            \bottomrule
              \\
          \end{tabular}
          }
        \label{tab-arch}
\end{table}

\subsection{Architecture Vectors and Architecture Embeddings}
We calculate architecture vectors $v_*$ and architecture embedding $h_*$ ($*\in\{G, D\}$) for our pruning agents as follows:

\begin{equation}
\begin{aligned}
y_*^l, h_*^l &= \textrm{GRU}(x_*^l, h_*^{l-1}), \\
{o}_*^l &= \textrm{Dense}_l(y_*^l), \\ l&=1,\cdots,L_*\ *\in\{G, D\},
\end{aligned}
\end{equation}

Take the discriminator $D$ as an example, we use let the last layer hidden outputs $h_D^{L_D}$ as the corresponding architecture embedding $h_D$. We let the initial hidden state $h_D^0 = h_G$. We concatenate all $o_*^l$ to get the final $o_D = [o_D^1,\cdots,o_D^{L_D}]$ used in Eq.~3 and Eq.~4. Similar procedures are applied for the generator $G$.

\begin{algorithm*}[!t]
    % \SetAlgoLined
    % \SetNoFillComment
    \small
\textbf{Input}: A pruning dataset of source domain images, their corresponding predictions in the target domain (by the original generator), and the set of neighbors for the predicted images on the original model's learned manifold (Section 3.3) $\mathcal{D}_p=\{(x_{i}), (y'_{i}, \mathcal{N}_{y'_i, k})\}_{i=1}^N$; Pruning agents $gen_G(\theta_G)$ and $gen_D(\theta_D)$; original generator and discriminator $G$ and $D$; Number of iterations $T$\;

\textbf{Output:} Pruning agents $gen_G(\theta_G)$ and $gen_D(\theta_D)$.\;

\textbf{Initialization:} Freeze the pretrained weights of $G$ and $D$ and disable gradient calculation for them.\;
\begin{algorithmic}
 \For{$t := 1$ to $T$}
    %%%%%%%%%%%%%%%%%%
    \State 1. Sample a pruning pair  $d = \{(x_{i}, (y'_i, \mathcal{N}_{y'_i, k}))\}$\;
    %%%%%%%%%%%%%%%%%%
    % \tcc{\textbf{****** Forward Pass of the Adversarial Game *****}}
    \State 2. $h_D \gets gen_D$
    ~~~\Comment{$h_D$ gets architecture embedding of $gen_D$ for $D$.}\;
    %%%%%%%%%%%%%%%%%%
    \State 3. $o_G, h_G \gets gen_G(h_D.detach(); \theta_G)$\;
    %%%%%%%%%%%%%%%%%%
    \State 4. $v_G \gets Gumbel$-$Sigmoid(o_G)$\;
    %%%%%%%%%%%%%%%%%%
    \State 5. Determine the architecture of $G$ using $v_G$.\;
    %%%%%%%%%%%%%%%%%%
    \State 6. $y^{fake} \gets G(x_i; v_G)$\;
    %%%%%%%%%%%%%%%%%% 
    % \tcc{\textbf{***************** Updating $gen_D$ *****************}}
    \State 7. $h_G \gets gen_G$\;
    % ~~~\tcp{$h_G$ gets architecture embedding of $gen_G$ for $G$.}
    %%%%%%%%%%%%%%%%%%
    \State 8. $o_D, h_D \gets gen_D(h_G.detach(); \theta_D)$
    \Comment{Do not backprop gradients for $gen_G$ when updating $gen_D$.}\;
    %%%%%%%%%%%%%%%%%%
    \smallskip
    \State 9. $v_D \gets Gumbel$-$Sigmoid(o_D)$\;
    %%%%%%%%%%%%%%%%%%
    \State 10. Determine the architecture of $D$ using $v_D$.\;
    %%%%%%%%%%%%%%%%%%
    \State 11. $p_{D}^{fake} \gets D(x_i, y^{fake}.detach(); v_D)$,~~~ $p_{D}^{real} \gets [D(x_i, y'_j; v_D)$ for $y'_j$ in $\mathcal{N}_{y'_i, k}]$\;
    %%%%%%%%%%%%%%%%%%
    \State 12. Calculate objective (2) using $p_{D}^{fake}$ and $p_{D}^{real}$ and backpropagate the loss gradients for the parameters of $gen_D$($\theta_D$) using STE~\cite{bengio2013estimating}.\;
    \State 13. Update $\theta_D$ using Adam optimizer.\;
    % \tcc{\textbf{***************** Updating $gen_G$ *****************}}
    \State 14. $h_D \gets gen_D$
    ~~~\Comment{$h_D$ gets architecture embedding of $gen_D$ for $D$.}\;
    %%%%%%%%%%%%%%%%%%
    \State 15. $o_G, h_G \gets gen_G(h_D.detach(); \theta_G)$
    \Comment{Do not backprop gradients for $gen_D$ when updating $gen_G$.}\;
    %%%%%%%%%%%%%%%%%%
    \smallskip
    \State 16. $v_G \gets Gumbel$-$Sigmoid(o_G)$\;
    %%%%%%%%%%%%%%%%%%
    \State 17. Determine the architecture of $G$ using $v_G$.\;
    %%%%%%%%%%%%%%%%%%
    \State 18. $p_{G}^{fake} \gets D(x_i, y^{fake}; v_D)$\;
    %%%%%%%%%%%%%%%%%%
    \State 19. Calculate objective (2) using $p_{G}^{fake}$ and backpropagate the loss gradients for the parameters of $gen_G$($\theta_G$) using STE~\cite{bengio2013estimating}.\;
    \State 20. Update $\theta_G$ using Adam optimizer.\;
    \EndFor
\end{algorithmic}
\textbf{return} $gen_G(\cdot)$, $gen_D(\cdot)$.
\caption{Our Proposed Pruning Scheme}
\label{alg-1}
\end{algorithm*}

\section{Experimental Details}
As mentioned in the paper, we mainly follow the setup of the previous works~\cite{zhang2022wavelet,li2021revisiting,li2020gan,li2022learning} in our experiments.~We elaborate on our hyperparameter setting for original models' training, pruning, and fine-tuning of them in the following. 
\subsection{Original Models' Training} \label{original-params}
 We use the original repository\footnotemark~for Pix2Pix~\cite{isola2017image} and CycleGAN~\cite{zhu2017unpaired} to train the original models. We use Adam optimizer~\cite{DBLP:journals/corr/KingmaB14} with parameters $(\beta_1, \beta_2)=(0.5, 0.999)$ to train the generator and discriminator for all models. The hyperparameter settings for each experiments is summarized in Tab.~\ref{original-params-tab}. Similar to the original training schemes~\cite{isola2017image,zhu2017unpaired}, we use constant learning rate for half of the training and linearly decay it to zero in the rest of it except for Pix2Pix on Edges2Shoes.~We also utilize Batch Normalization~\cite{ioffe2015batch} and Instance Normalization~\cite{ulyanov2016instance} in our experiments for the architectures of Pix2Pix and CycleGAN respectively.
\footnotetext{\href{https://github.com/junyanz/pytorch-CycleGAN-and-pix2pix}{https://github.com/junyanz/pytorch-CycleGAN-and-pix2pix}}

\begin{table}[hbt!]
\caption{\small Hyperparameter settings for training original models.}
\label{original-params-tab}
\resizebox{\linewidth}{!}{
\begin{tabular}{c|c|c|c|cc|cc|c}
\hline
\multirow{2}{*}{Model} & \multirow{2}{*}{Dataset} & \multirow{2}{*}{GAN Loss} & \multirow{2}{*}{Batch Size} & \multicolumn{2}{c|}{Training Epochs}     & \multicolumn{2}{c|}{Optimization Params}   & \multirow{2}{*}{Normalization} \\ \cline{5-8} & & & & \multicolumn{1}{c|}{Constant LR} & Decay  & \multicolumn{1}{c|}{LR} & Weight Decay & \\ \hline
Pix2Pix & Cityscapes & hinge & 1 & \multicolumn{1}{c|}{100} & 100 & \multicolumn{1}{c|}{0.0002} & 0 & Batch \\ \hline
Pix2Pix & Edges2Shoes & hinge & 4 & \multicolumn{1}{c|}{5} & 25 & \multicolumn{1}{c|}{0.0002} & 0 & Batch \\ \hline
CycleGAN & Horse2Zebra & Least Squares & 1 & \multicolumn{1}{c|}{100} & 100 & \multicolumn{1}{c|}{0.0002} & 0 & Instance \\ \hline
CycleGAN & Summer2Winter & Least Squares & 1 & \multicolumn{1}{c|}{100} & 100   & \multicolumn{1}{c|}{0.0002} & 0  & Instance \\ \hline
\end{tabular}}
\end{table}

\subsection{Pruning}
We prune the original pretrained GAN models using our proposed Obj.~2 in the paper and two pruning agents with the architectures described in section Pruning. We set $\lambda_1=3.0$, $\lambda_2=0.1$, and the number of neighbor samples $k=5$ during pruning for all the datasets. We also use Adam~\cite{DBLP:journals/corr/KingmaB14} with parameters $(\beta_1, \beta_2)=(0.9, 0.999)$ for pruning. Other pruning hyperparameters are summarized in Tab.~\ref{pruning-params}.~For the self-supervised SwAV~\cite{caron2020unsupervised} model, we take the provided checkpoint for ResNet-50, trained on ImageNet for 800 epochs, in their original\footnotemark~repository.~Then, we fine-tuned the models with batch size of 256 for 10 epochs.~We set the number of clusters to $100$ for Cityscapes, $200$ for Edges2Shoes, and $50$ for Horse2Zebra as well as Summer2Winter. We set the other parameters the same as the default ones used in the SwAV~\cite{caron2020unsupervised} training from scratch. We provide our pruning algorithm in alg.~\ref{alg-1}.

\footnotetext{\href{https://github.com/facebookresearch/swav}{https://github.com/facebookresearch/swav}}

\begin{table}[hbt!]
\caption{\small Hyperparameter settings for pruning agents.}
\label{pruning-params}
\resizebox{\linewidth}{!}{
\begin{tabular}{c|c|c|c|c|cc}
\hline
\multirow{2}{*}{Model} & \multirow{2}{*}{Dataset} & \multirow{2}{*}{\begin{tabular}[c]{@{}c@{}}Manifold\\ Loss\end{tabular}} & \multirow{2}{*}{Batch Size} & \multirow{2}{*}{Pruning Epochs} & \multicolumn{2}{c}{Optimization Params}    \\ \cline{6-7} 
&                          &                                                                          &                             &                                 & \multicolumn{1}{c|}{LR}     & Weight Decay \\ \hline
Pix2Pix                & Cityscapes               & hinge                                                                    & 1                           & 20                              & \multicolumn{1}{c|}{0.001} & 0.0001        \\ \hline
Pix2Pix                & Edges2Shoes              & hinge                                                                    & 4                           & 3                               & \multicolumn{1}{c|}{0.001} & 0.0001        \\ \hline
CycleGAN               & Horse2Zebra              & Least Squares                                                            & 1                           & 20                              & \multicolumn{1}{c|}{0.001} & 0.0001        \\ \hline
CycleGAN               & Summer2Winter            & Least Squares                                                            & 1                           & 20                              & \multicolumn{1}{c|}{0.001} & 0.0001        \\ \hline
\end{tabular}}
\end{table}

\subsection{Fine-tuning}
After pruning stage, we employ the trained pruning agents to determine the optimal sub-structure of the generator and discriminator. Then, we use the original loss functions for Pix2Pix~\cite{isola2017image} and CycleGAN~\cite{zhu2017unpaired} to fine-tune the models with Adam optimizer~\cite{DBLP:journals/corr/KingmaB14}. Thus, most of the hyperparameters are similar to section~\ref{original-params} except a few changes. 
% Firstly, we use Adam optimizer~\cite{DBLP:journals/corr/KingmaB14} with parameters $(\beta_1, \beta_2)=(0.5, 0.999)$. 
We use knowledge distillation~\cite{hinton2015distilling,li2021revisiting} between the original generator and the pruned one. However, as mentioned in the paper, we do not use online distillation in our model, which saves computational costs. We use the Perceptual loss~\cite{johnson2016perceptual} for distillation that consists of two components: 1) Content loss that is the squared norm of the difference between two feature maps of the original and pruned models. 2) Texture loss that is the Frobenius norm of the difference between their Gram matrices. These two losses are applied on the same layers as GCC~\cite{li2021revisiting}. We represent their coefficients in our loss function with $\lambda_{content}$ and $\lambda_{texture}$ respectively. Tab.~\ref{fine-tune-params} shows our hyperparameter setting for fine-tuning the models in each experiment.

\begin{table}[hbt!]
\caption{\small Hyperparameter settings for Fine-tuning.}
\resizebox{\linewidth}{!}{
\begin{tabular}{c|c|c|c|c|cc|c|c}
\hline
\multirow{2}{*}{Model} & \multirow{2}{*}{Dataset} & \multirow{2}{*}{\begin{tabular}[c]{@{}c@{}}GAN\\ Loss\end{tabular}} & \multirow{2}{*}{Batch Size} & \multirow{2}{*}{\begin{tabular}[c]{@{}c@{}}Fine-tuning\\ Epochs\end{tabular}} & \multicolumn{2}{c|}{Optimization Params}   & \multirow{2}{*}{$\lambda_{content}$} & \multirow{2}{*}{$\lambda_{texture}$} \\ \cline{6-7}
    &  &  &  &  & \multicolumn{1}{c|}{LR} & Weight Decay & &    \\ \hline Pix2Pix                & Cityscapes               & hinge                                                               & 1                           & 50                                                                            & \multicolumn{1}{c|}{0.0002} & 0         & 50                               & 10000                         \\ \hline
Pix2Pix                & Edges2Shoes              & hinge                                                               & 4                           & 50                                                                            & \multicolumn{1}{c|}{0.0002} & 0         & 50                               & 10000                         \\ \hline
CycleGAN               & Horse2Zebra              & Least Squares                                                       & 1                           & 50                                                                            & \multicolumn{1}{c|}{0.0002} & 0       & 0.01                             & 0                            \\ \hline
CycleGAN               & Summer2Winter            & Least Squares                                                       & 1                           & 50                                                                            & \multicolumn{1}{c|}{0.0002} & 0        & 0.01                             & 0                             \\ \hline
\end{tabular}}
\label{fine-tune-params}
\end{table}

\end{document}